\pdfoutput=1

\documentclass[11pt]{article}

\usepackage[preprint]{acl}

\usepackage{times}
\usepackage{latexsym}
\usepackage{amsfonts}
\usepackage{xcolor}
\usepackage{subcaption}

\definecolor{preprocessing_color}{RGB}{248, 186, 100} 
\definecolor{feature_color}{RGB}{71, 159, 248} 
\definecolor{analysis_color}{RGB}{129, 214, 83} 
\definecolor{toolkit_color}{RGB}{235, 235, 235}

\usepackage[T1]{fontenc}

\usepackage[utf8]{inputenc}

\usepackage{microtype}

\usepackage{inconsolata}
\usepackage{graphicx}

\def\toolname{\texttt{Edu-ConvoKit}}

\newcommand{\icon}{
    \includegraphics[scale=0.4]{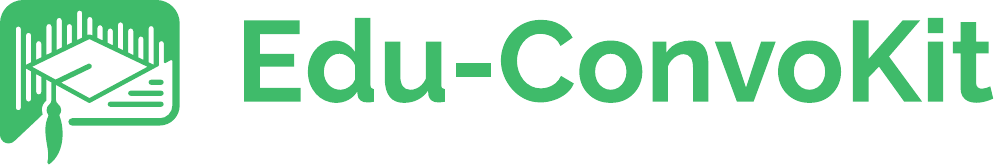}}

\newcommand{\miniicon}{
    \includegraphics[scale=0.02]{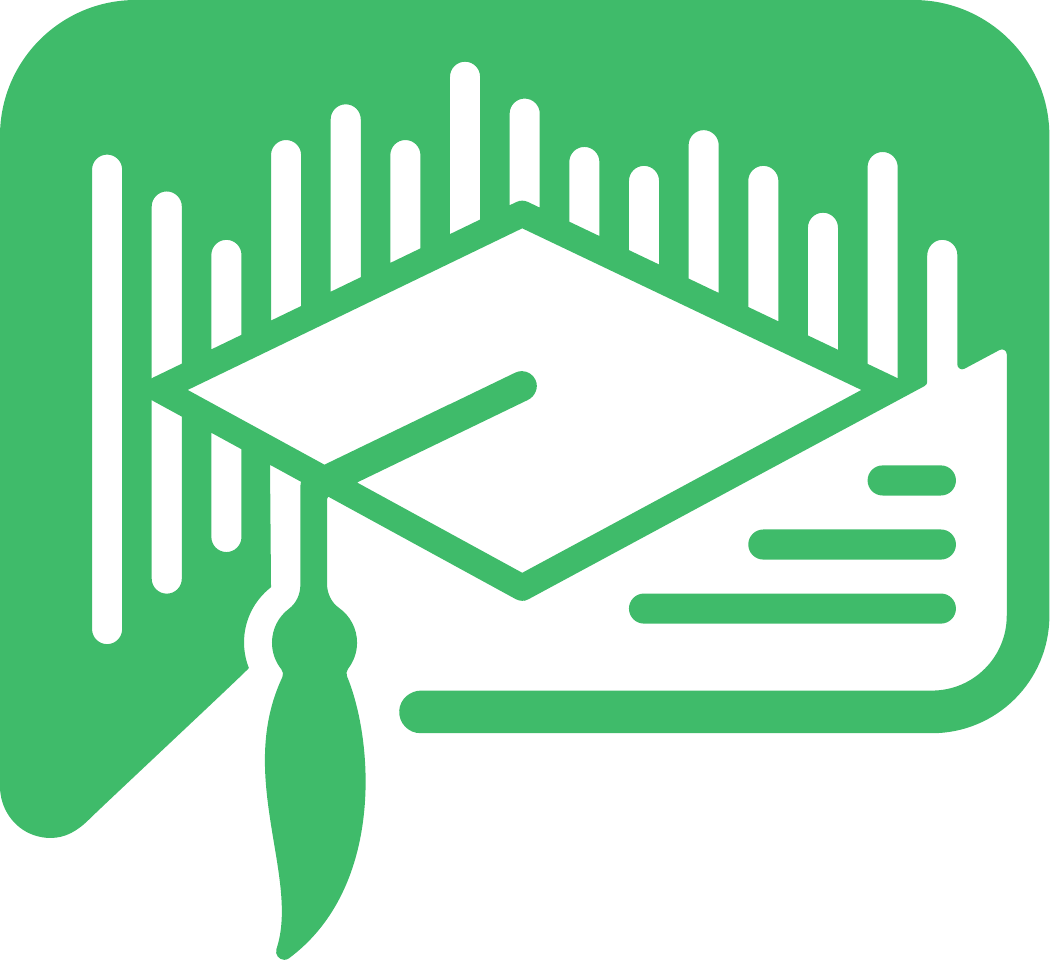}}

\newcommand{\colabicon}{
    \includegraphics[height=1em]{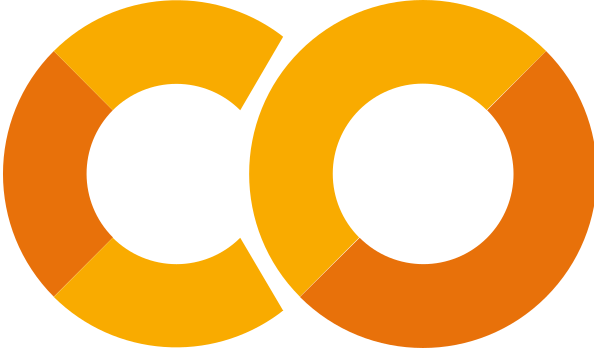}}
    
\newcommand{\papersicon}{
    \includegraphics[height=1em]{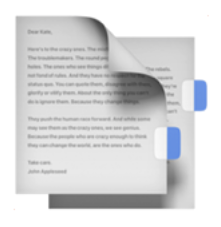}}
    
\title{\icon{}\\
An Open-Source Library for Education Conversation Data
}

\author{Rose E. Wang \\
  Stanford University\\
  \texttt{rewang@cs.stanford.edu} \\\And
  Dorottya Demszky \\
  Stanford University \\
  \texttt{ddemszky@stanford.edu} \\}

\begin{document}
\maketitle
\begin{abstract}
We introduce\miniicon{} \toolname{}, an open-source library designed to handle pre-processing, annotation and analysis of conversation data in education.
Resources for analyzing education conversation data are scarce, making the research challenging to perform and therefore hard to access.
We address these challenges with \toolname{}.
\toolname{} is open-source\footnote{\url{https://github.com/stanfordnlp/edu-convokit}}, pip-installable\footnote{\url{https://pypi.org/project/edu-convokit/}}, with comprehensive documentation\footnote{\url{https://edu-convokit.readthedocs.io/en/latest/}}.
Our demo video is available at: \url{https://youtu.be/zdcI839vAko?si=h9qlnl76ucSuXb8-}.
We include additional resources, such as\colabicon{} Colab applications of \toolname{} to three diverse education datasets\footnote{\url{https://github.com/stanfordnlp/edu-convokit?tab=readme-ov-file\#datasets-with-edu-convokit}} and a\papersicon{} repository of \toolname{}-related papers\footnote{\url{https://github.com/stanfordnlp/edu-convokit/blob/main/papers.md}}.
\end{abstract}

\section{Introduction}
Language is central to educational interactions, ranging from classroom instruction to tutoring sessions to peer discussions. 
It offers rich insights into the teaching and learning process that go beyond the current, oversimplified view of relying on standardized test outcomes \citep{wentzel1997student,pianta2003relationships,robinson2022framework, wentzel2022does}. 
The landscape of natural language processing (NLP) and education is rapidly evolving, with an increase of open-sourced education conversation datasets (e.g., from \citet{caines2020teacher, stasaski-etal-2020-cima, SureshJacobsLaiTanWardMartinSumner2021, demszky-hill-2023-ncte, wang2023sight, wang2023stepbystep, holt2023amber}), heightened interest manifesting in academic venues (e.g., NeurIPS \citet{neurips2023gaied}, Building Educational Applications at $^*$ACL Conferences \citet{acl2023bea}, and education conferences hosting NLP tracks\footnote{The International Conference on Learning Analytics and Knowledge (LAK), Education Data Mining (EDM), and Artificial Intelligence in Education (AIED).}), 
alongside courses dedicated to this field (e.g., Stanford's NLP and Education course CS293\footnote{\url{https://web.stanford.edu/class/cs293/}}).

\begin{figure*}
    \centering
    \includegraphics[width=\textwidth]{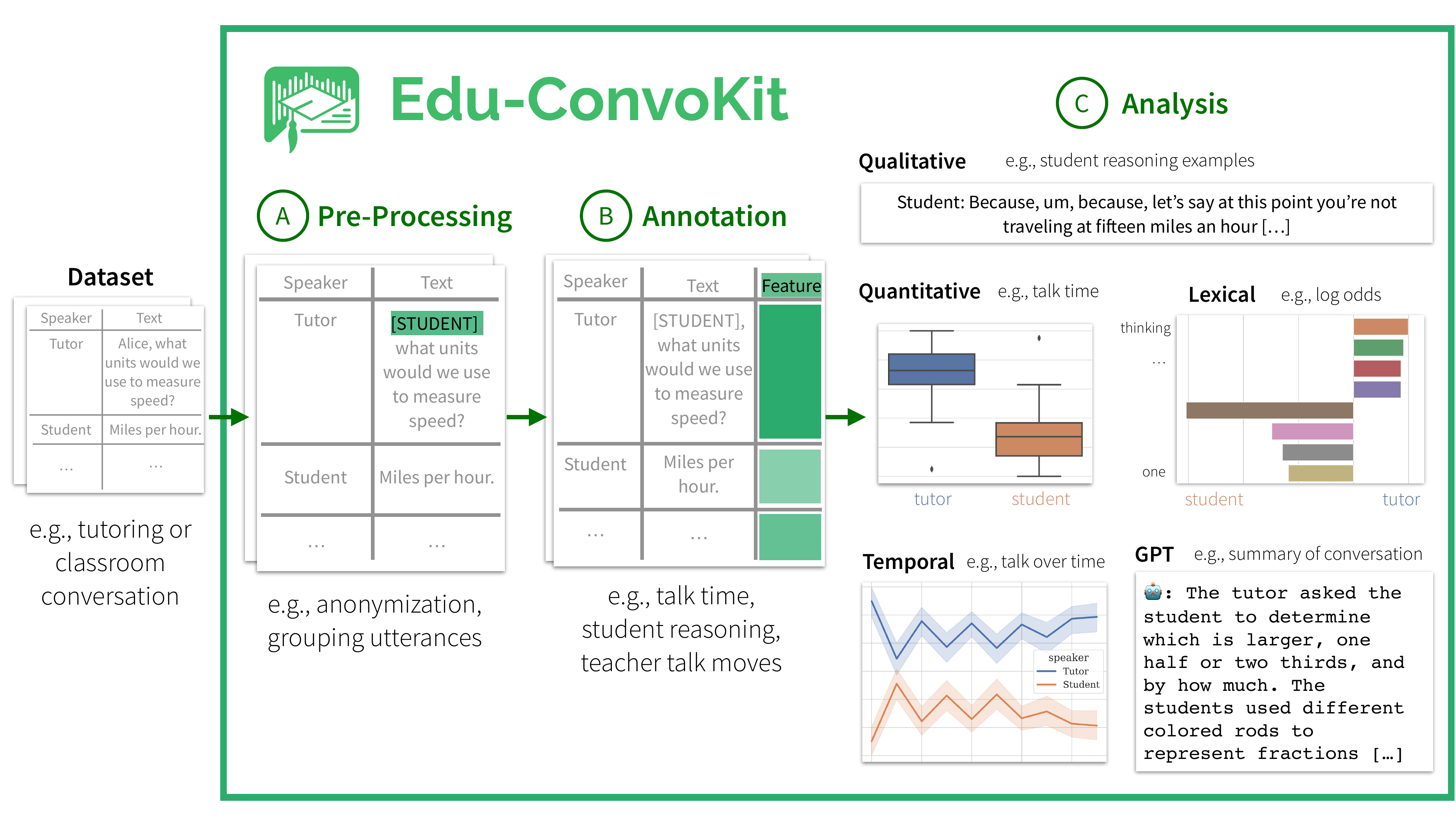}
    \caption{\textbf{Overview of\miniicon{} \toolname.} \toolname{} is designed to facilitate the study of conversation data in education. 
    It is a modular, end-to-end pipeline for \textbf{A.} pre-processing, \textbf{B.} annotating, and \textbf{C.} analyzing education conversation data.
    As additional resources, the toolkit includes\colabicon{} Colab notebooks applying \toolname{} to three existing, large education datasets and a centralized database of \toolname{} \papersicon{} papers.
    This toolkit aims to enhance the accessibility and reproducibility of NLP and education research.
    }
    \label{fig:main_figure}
\end{figure*}

\paragraph{Challenges and consequences.}
While the interest in this interdisciplinary field is growing, our conversations with education data science and NLP researchers both in academia and industry have surfaced several challenges that hinder research progress. 
First, there is \textbf{no centralized tool or resource} that assists in analyzing education data, or helps researchers understand different  tradeoffs in methods.
For example, researchers expressed uncertainty about pre-processing the data, such as ``the best way to anonymize the data to protect the privacy of students and teachers''. 
They also wanted an ``easily accessible collection of language tools and models that can detect insightful things.'' 
The lack of these tools and resources makes the research harder to conduct.
Second, there is a \textbf{high learning curve for performing computational analyses}.
For example, many education researchers are trained in qualitative research; even though they want to use computational tools for quantitative analyses at scale, they often do not know how to start or have the readily available compute to try out the tools.

\paragraph{Our system.} Our work introduces\miniicon{}\toolname{} to address these challenges.
\toolname{} is designed to facilitate and democratize the study of education conversation data. 
It is a modular, end-to-end pipeline for \textbf{A.} pre-processing, \textbf{B.} annotating, and \textbf{C.} analyzing education conversation data, illustrated in Figure~\ref{fig:main_figure}. 
Specifically, \toolname{}
\begin{itemize}
    \item \textbf{Supports pre-processing} for education conversation datasets, such as automatically de-identifying conversations; 
    \item \textbf{Hosts a collection of language tools and models for annotation}, ranging from traditional (e.g., talk time) to neural measures (e.g., classifying student reasoning); and
    \item \textbf{Automates several analyses} used in NLP and education research, ranging from qualitative analyses, temporal analyses and GPT-powered analyses (e.g., on summarizing transcripts). 
\end{itemize}

To demonstrate its flexible design and ensure its accessibility regardless of compute infrastructure, we created\colabicon{} \textbf{Colab notebooks of \toolname{} applied to three diverse education conversation datasets} in mathematics \citep{demszky-hill-2023-ncte, suresh2021using, holt2023amber}. 
We additionally created \textbf{a centralized database of research projects} that have either used \toolname{} or have features integrated in the toolkit. 
We invite the community to contribute to the toolkit and collectively push the boundaries of education conversation research!

\section{Related Works}

\subsection{Advancing NLP through Toolkits}
The NLP  community has benefited greatly from the public availability of general toolkits, which standardize the way data is transformed, annotated and analyzed.
Examples include NLTK \citep{bird2006nltk}, StanfordNLP \citep{qi2019universal}, spaCy \citep{honnibal2020spacy}, or scikit-learn \citep{pedregosa2011scikit}. 
They improve the accessibility to the research and allow researchers to focus on developing new methods, rather than on re-implementing existing ones.
\toolname{} shares these goals. 
ConvoKit \citep{chang2020convokit} is a NLP package for conversational analysis and bears the most similarity to our work.
A key difference between our library and ConvoKit is the data structure: \toolname{} uses a table-based dataframe structure whereas ConvoKit uses an object-based data structure akin to a dictionary.
Our data structure makes manipulating data easier, e.g., performing utterance-level annotations. 
Additionally, our tool caters to education language research and therefore supports an array of common analyses such as qualitative analysis \citep{erickson1985qualitative, corbin1990grounded, wang-etal-2023-sight}, quantitative evaluations \citep{bienkowski2012enhancing, kim2023high, demszky2023improving}, or lexical comparisons \citep{praharaj2021towards, handa2023mistakes}. 

\subsection{Supporting the Multifaceted Nature of Education Interaction Research}

\toolname{} sits at the intersection of many disciplines that use different annotation and analysis tools for understanding language use in education interactions. 
For example, qualitative education research uses \textit{qualitative analysis} to manually analyze the discourse, such as how students collaborate with each other \citep{mercer1996quality, jackson2013exploring, langer2020so, chen2020visual, hunkins2022beautiful}.
Learning analytics uses \textit{quantitative and temporal analysis} to summarize statistics in aggregate or over time  \citep{bienkowski2012enhancing, kim2023high, demszky2023improving, demszky2024does}. 
Other areas perform \textit{lexical analyses and neural measures for annotating} education discourse features  \citep{reilly2019predicting, praharaj2021towards, rahimi2017assessing, alic-etal-2022-computationally, hunkins2022beautiful, demszky-hill-2023-ncte, reitman2023multi, SureshJacobsLaiTanWardMartinSumner2021,zachary2023a, wang-demszky-2023-chatgpt}. 
Recently, newer analysis tools powered by GPT models analyze complete conversations such as summarizing or pulling good examples of teacher instruction from the classroom transcripts \citep{wang-demszky-2023-chatgpt}.
\toolname{} is designed to support these forms of annotation and analysis, and unify the currently fragmented software ecosystem of this interdisciplinary research area.

\section{Design Principles}
\toolname{} follows these principles:

\begin{description}
    \item[I.] \textbf{Minimalistic Data Structure}. The system transforms all data inputs (e.g., csv and json files) into a dataframe. 
    \toolname{} only needs the speaker and text columns to be uniquely identifiable, which is the case in the datasets we surveyed and applied \toolname{} to. 
    \item[II.] \textbf{Efficient Execution}. The system should be able to run on a CPU and support large-scale pre-processing, annotation and analysis. 
    \item[III.] \textbf{Modularity}. Each component of \toolname{} functions as an independent module. 
    Running one module (e.g., pre-processing) should not be required for the user to run another module (e.g., annotation).
\end{description}

These principles enable \toolname{} to comprehensively incorporate different methods for pre-processing, annotation and analysis. 
They ensure that \toolname{} is effective and adaptable to various research needs.

\section{\includegraphics[height=2em]{images/fitted_ECK-Lockup-Green.pdf}}

\toolname{} is organized around three entities: 
\texttt{PreProcessor}, \texttt{Annotator}, and \texttt{Analyzer} (see Figure~\ref{fig:main_figure}).
The following sections enumerate each entity's functionality. 
Please refer to the short demo video to preview \toolname{} in action: \url{https://youtu.be/zdcI839vAko?si=h9qlnl76ucSuXb8-}.

\subsection{\texttt{PreProcessor}\label{sec:preprocessor}} 

The \texttt{PreProcessor} module in \toolname{} processes the raw data and includes several techniques standard to education and NLP research practices, such as replacing speaker names with unique identifiers, merging consecutive utterances by the same speaker, and formatting text to be human-readable. 
Figure~\ref{fig:annotation_example} illustrates a simple example of text de-identification with \texttt{PreProcessor}, assuming that the researcher has access to a list of names (e.g. classroom roster) to be replaced.
\texttt{PreProcessor} accounts for multiple names per individual, and users can define how each name should be replaced. 
This feature ensures that the context of each interaction is preserved while maintaining confidentiality of the participants.

\begin{figure}[h!]
    \centering
    \includegraphics[width=\linewidth]{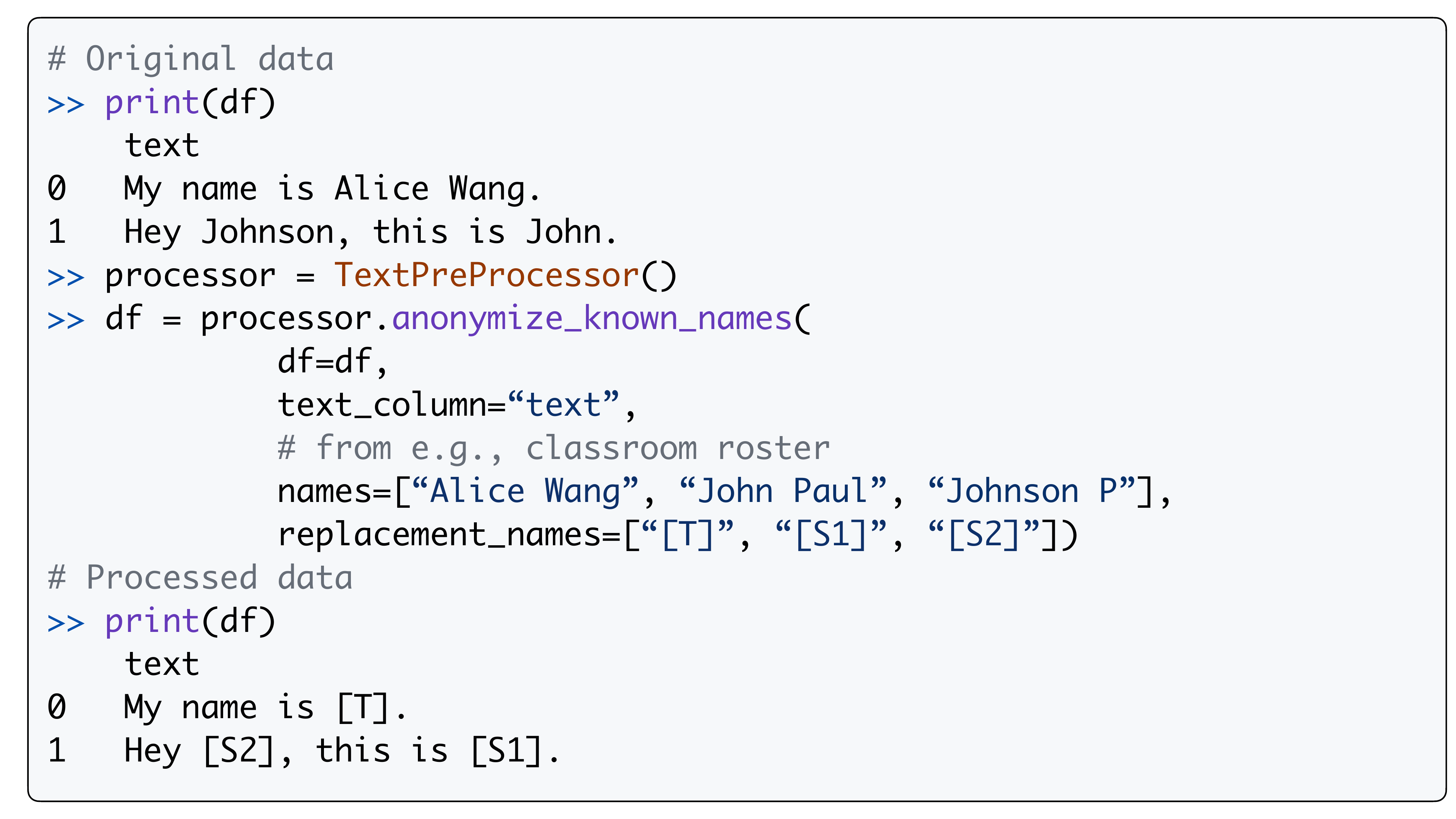}
    \caption{\textbf{Example for text de-identification.} \texttt{PreProcessor} accounts for multiple names (e.g., ``John Paul'' matches to ``John''), and handles word boundaries (e.g., ``John'' does not match to ``Johnson'').}
    \label{fig:anonymization_example}
\end{figure}

\subsection{\texttt{Annotator} \label{sec:annotation}}

\begin{figure}[h!]
    \centering
    \includegraphics[width=\linewidth]{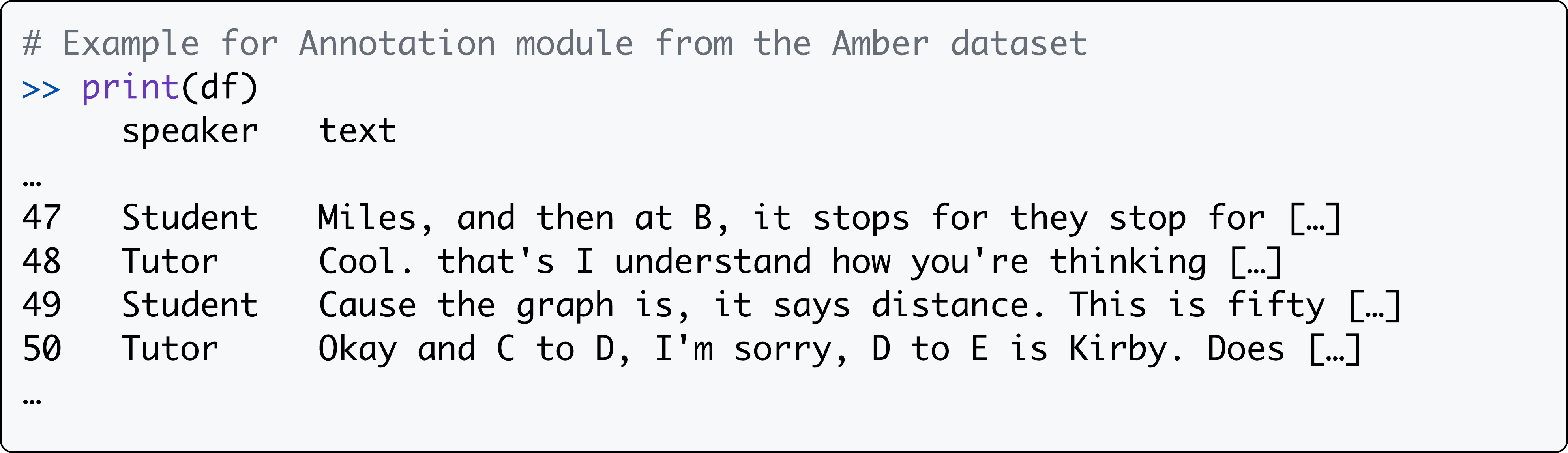}
    \caption{\textbf{Example for \texttt{Annotator}}.}
    \label{fig:annotation_example}
\end{figure}

\texttt{Annotator} annotates features at an \textit{utterance-level}.
It currently supports 7 types of features, ranging from traditional to neural measures of educational discourse.
The features follow the original implementations of cited works and the neural measures are models hosted on HuggingFace hub.
Notably, \texttt{Annotator} performs annotation with a single function call.
The following sections describe these features, using  Figure~\ref{fig:annotation_example} as the running example.

\paragraph{Talk Time.} Talk time measures the amount of speaker talk by word count and timestamps (if provided in the dataset). 
This feature quantifies the participation of both teachers/tutors and students, offering insights into classroom dynamics \citep{teachfx, jensen2020toward, demszky2024does}.

\begin{figure}[h!]
    \centering
    \includegraphics[width=\linewidth]{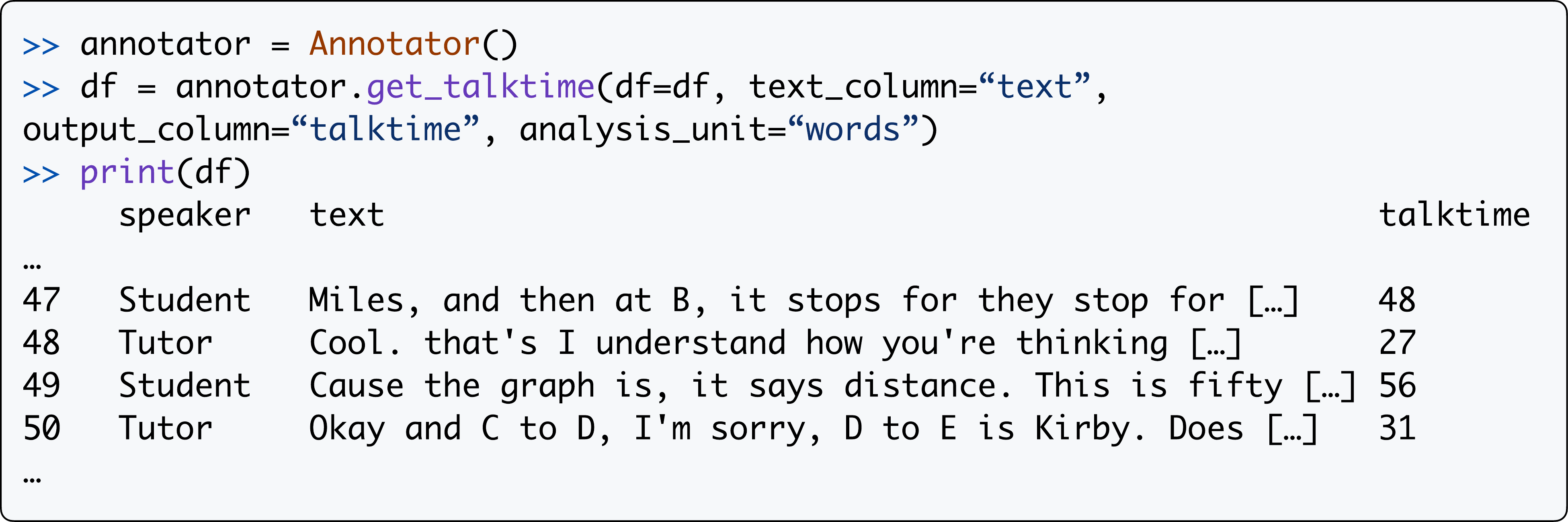}
\end{figure}

\paragraph{Math Density.} 
Math density measures the number of math terms used in an utterance, where the dictionary of math terms was collected in prior work by mathematics education researchers \citep{zachary2023a}.
This feature provides a quantitative measure of mathematical content in the dialogue.

\begin{figure}[h!]
    \centering
    \includegraphics[width=\linewidth]{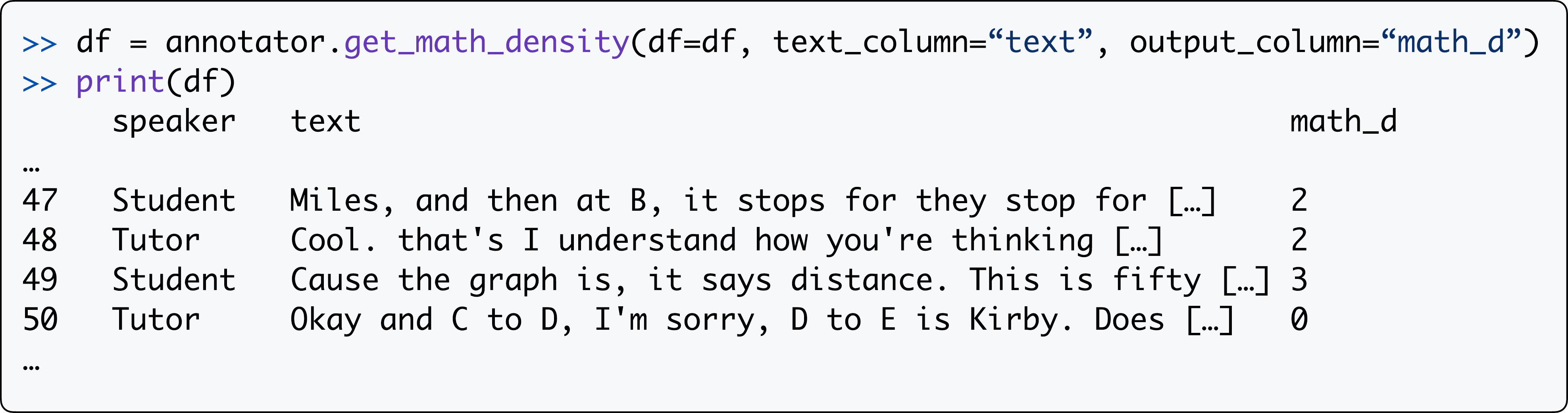}
\end{figure}

\paragraph{Student Reasoning.} 
The student reasoning annotation measures whether a given student utterance provides a mathematical explanation for an idea, procedure or solution \citep{demszky-hill-2023-ncte,hill2008mathematical}.
The model is a finetuned RoBERTa classifier \citep{liu2019roberta} on instances of student reasoning from elementary math classroom transcripts.
\toolname{} follows the original implementation from \citet{demszky-hill-2023-ncte}, ensuring fidelity to prior research: \texttt{Annotator} only label utterances that are at least 8 words long based on word boundaries; all other utterances are annotated as NaN. 
Furthermore, users can also easily specify which speakers to annotate for, such as to only annotate the student speakers as shown in the example below.

\begin{figure}[h!]
    \centering
    \includegraphics[width=\linewidth]{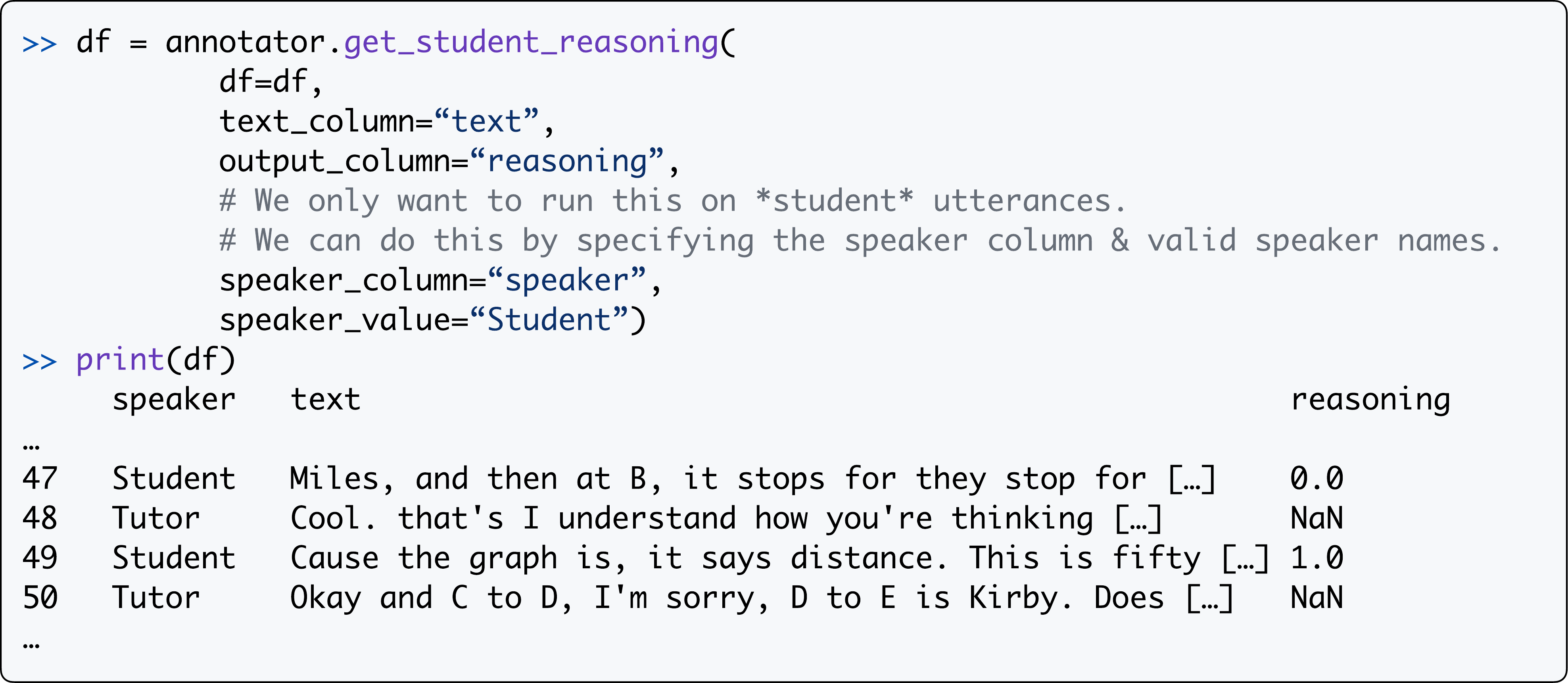}
\end{figure}

\paragraph{Focusing Questions.} 
The focusing question annotation capture questions that attend to what the student is thinking and presses them to communicate their thoughts clearly \citep{leinwarnd2014national, alic-etal-2022-computationally}. 
The model is a finetuned RoBERTa classifier \citep{liu2019roberta} on instances of teacher focusing questions from elementary math classroom transcripts: 

\begin{figure}[h!]
    \centering
    \includegraphics[width=\linewidth]{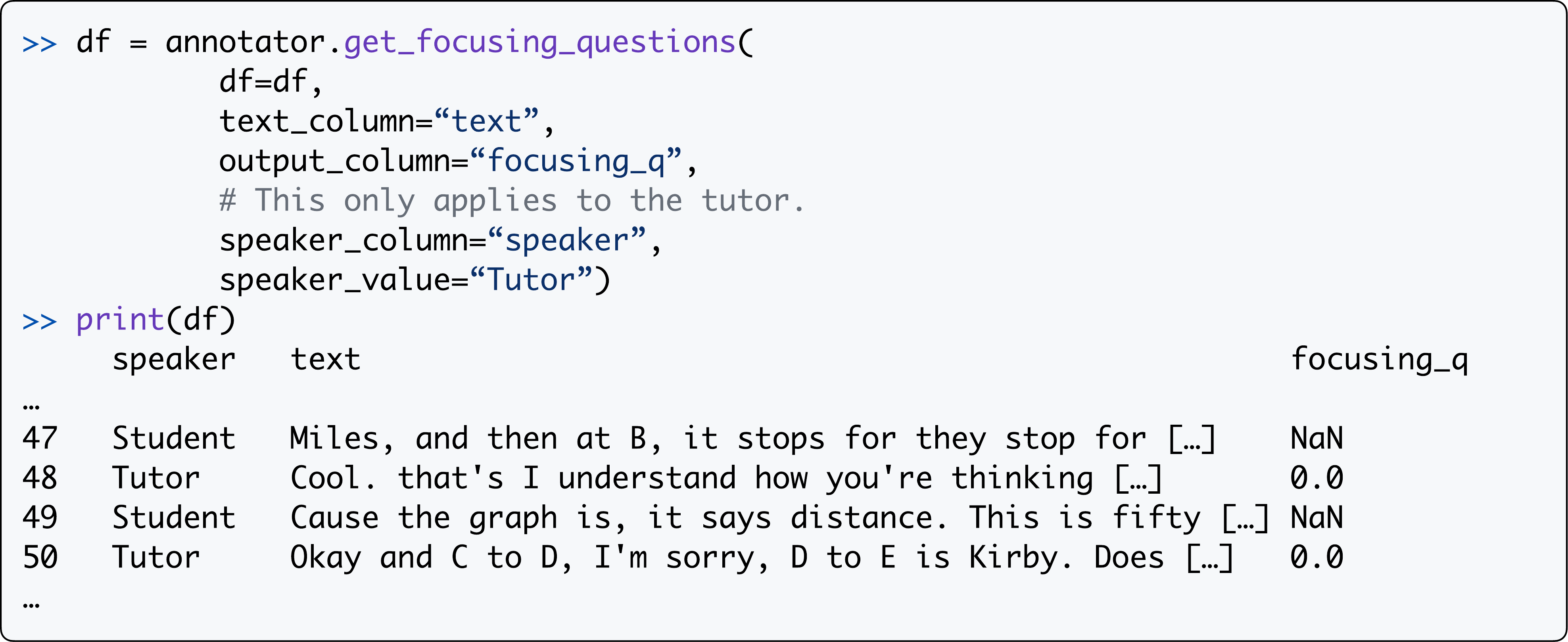}
\end{figure}

\paragraph{Teacher Accountable Talk Moves.} 
Teacher accountable talk moves capture the teacher's strategies to promote equitable participation in classrooms \citep{suresh2021using, jacobs2022promoting}, based on the Accountable Talk framework \citep{connor_2015_scaling}. 
It is a finetuned ELECTRA 7-way classifier \citep{clark2020electra} where: 0: No Talk Move Detected, 1: Keeping Everyone Together, 2: Getting Students to Related to Another Student's Idea, 3: Restating, 4: Revoicing, 5: Pressing for Accuracy, 6: Pressing for Reasoning.

\begin{figure}[h!]
    \centering
    \includegraphics[width=\linewidth]{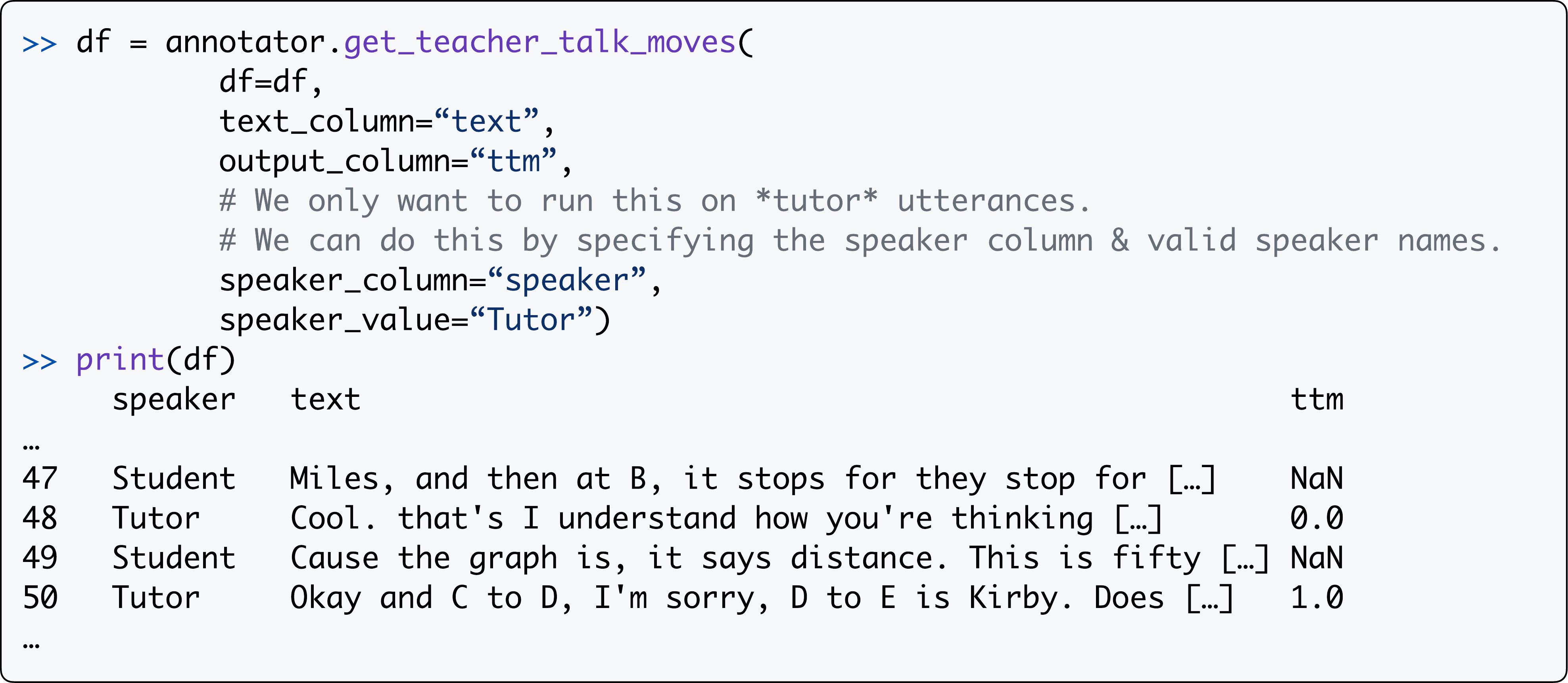}
\end{figure}

\paragraph{Student Accountable Talk Moves.} 
Analogous to the teacher talk moves, the student accountable talk moves are student discussion strategies to promote equitable participation in a rigorous classroom learning environment \citep{suresh2021using, jacobs2022promoting}. 
It is also a finetuned ELECTRA classifier for 5 classes: 0: No Talk Move Detected, 1: Relating to Another Student, 2: Asking for More Information, 3: Making a Claim, 4: Providing Evidence or Reasoning.

\begin{figure}[h!]
    \centering
    \includegraphics[width=\linewidth]{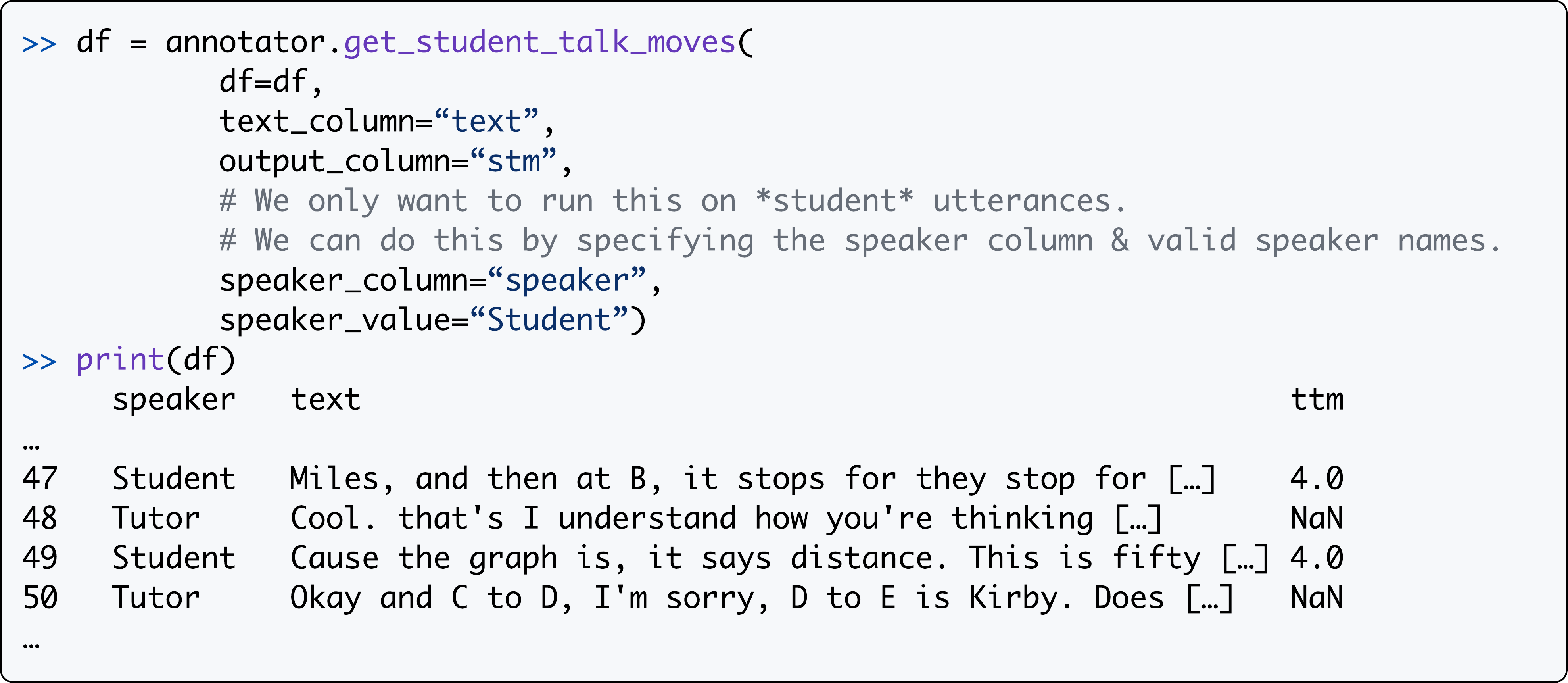}
\end{figure}

\paragraph{Conversational Uptake.}
Conversational uptake measures how teachers build on the contributions of students \citep{demszky-etal-2021-measuring}.
It is a BERT model fine-tuned with a self-supervised training objective (next utterance prediction), on an elementary math classroom dataset~\citep{demszky-hill-2023-ncte}, Switchboard~\citep{godfrey1997switchboard} and a tutoring dataset.
\texttt{Annotator} annotates utterances according to the original implementation: It can label teacher utterances following substantive student utterances that are at least 5 words long, such as in the example below.

\begin{figure}[h!]
    \centering
    \includegraphics[width=\linewidth]{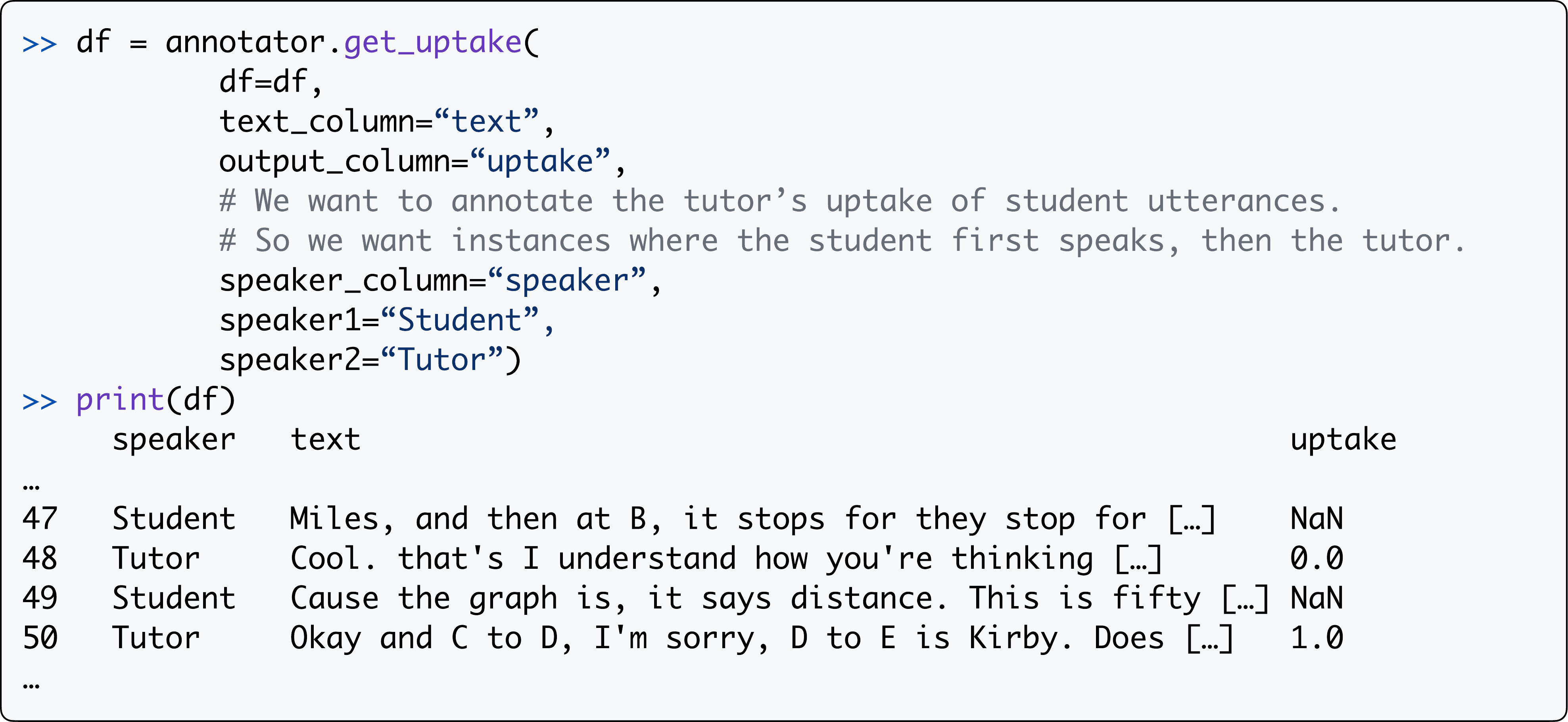}
\end{figure}

\subsection{\texttt{Analyzer}}

\toolname{} supports several modules that cover common analyses in education conversation research.
In general, each module is exposed by three methods: \texttt{plot} for plotting, \texttt{print} for displaying results in the terminal, and \texttt{report} for outputting results as text.
There are multiple data entry points for the \texttt{Analyzer} such as a single or multiple transcripts, or a data directory.
The following sections describe these modules, assuming that the variable \texttt{DATA\_DIR} is a directory of annotated transcripts.

\paragraph{\texttt{QualitativeAnalyzer}.}
This module enables researchers to view annotation examples. 
For example, we can easily view positive examples of student reasoning below.
This module has other features, such as additionally showing the previous and subsequent lines around the examples; please refer to our documentation for all features.

\begin{figure}[h!]
    \centering
    \includegraphics[width=\linewidth]{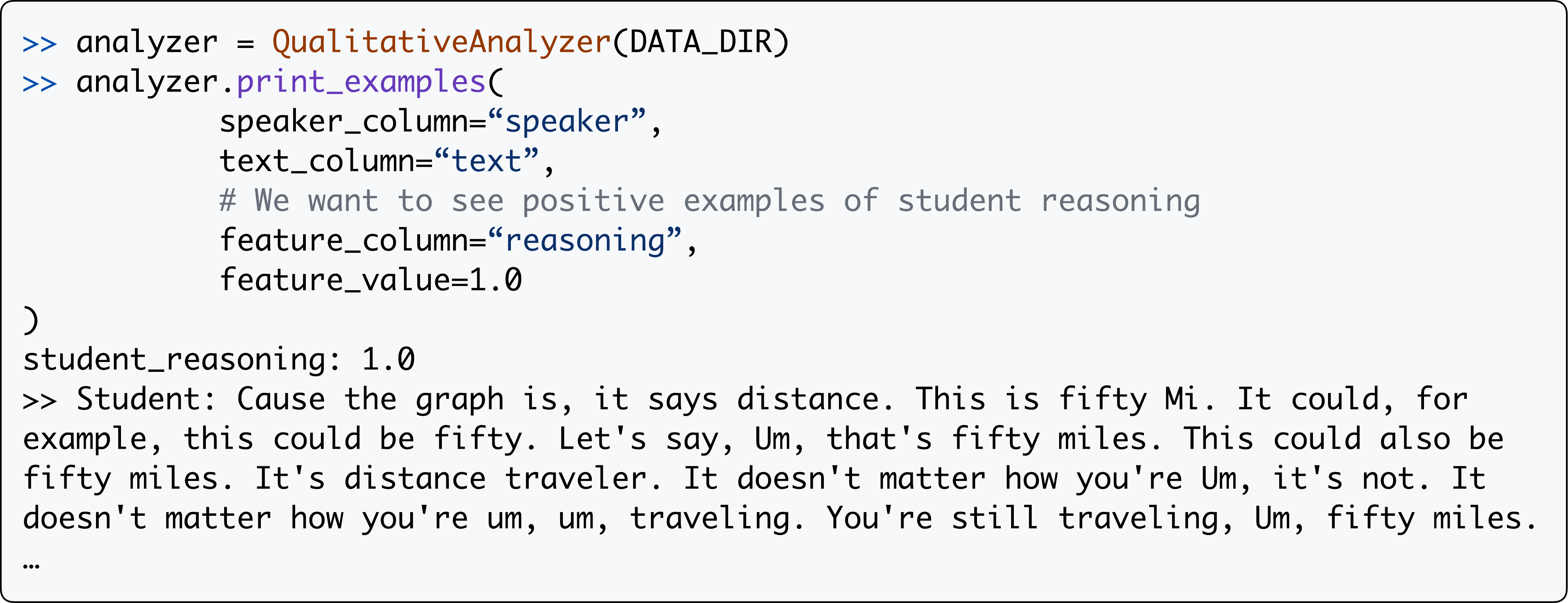}
    \label{fig:qualitative_example}
\end{figure}

\paragraph{\texttt{QuantitativeAnalyzer}.}
This module reports the quantitative summaries of the annotation results. 
Users can also flexibly group and use different representations, such as grouping by speaker or displaying the values as percentages as shown below.
\vspace{1em}
\begin{figure}[h!]
    \centering
    \includegraphics[width=\linewidth]{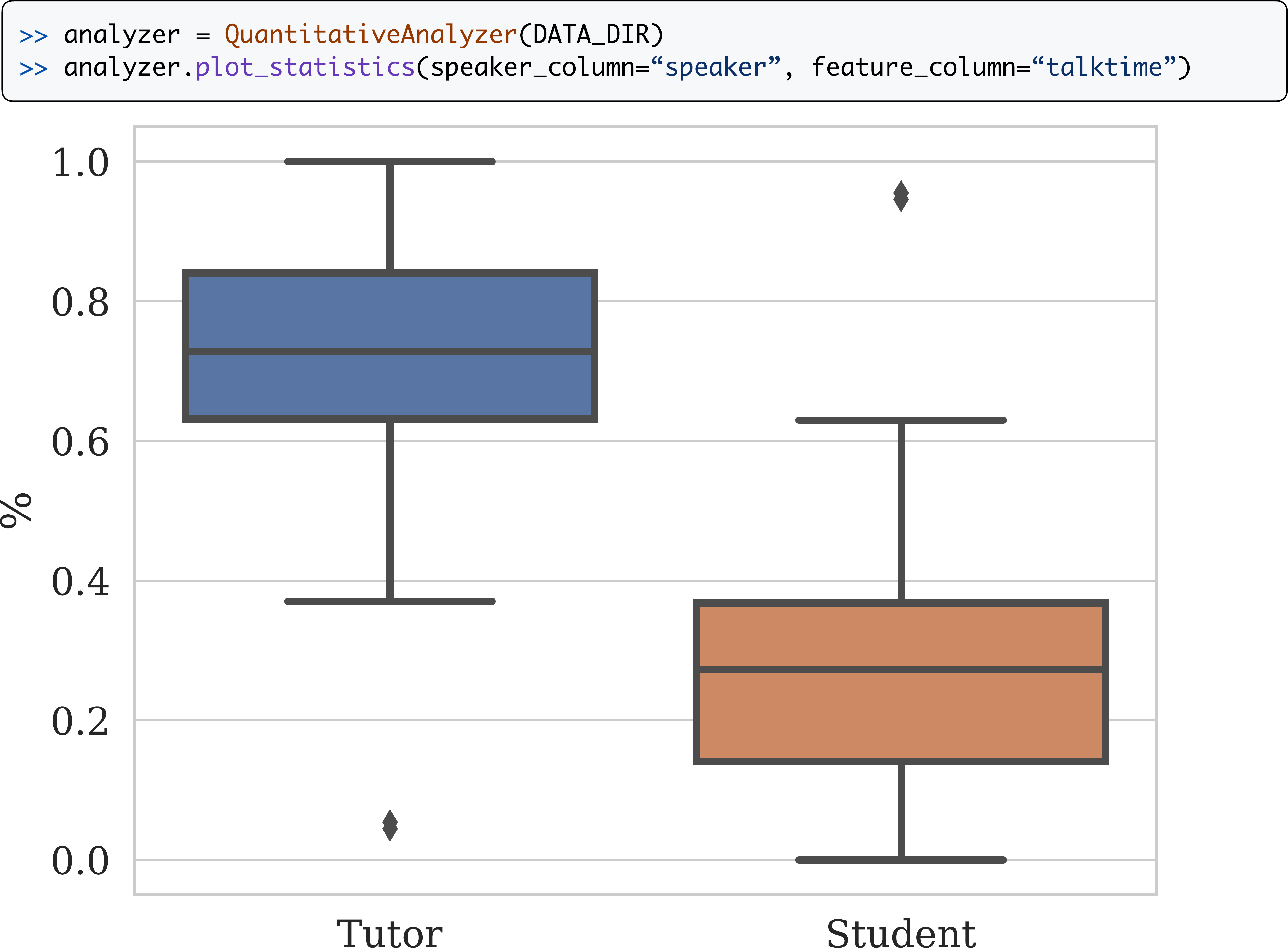}
\end{figure}

\paragraph{\texttt{LexicalAnalyzer}.}
This module reports language patterns on the word-level. 
It can report n-gram frequency and weighted log-odds analysis from Section 3.4 of \citet{monroe2008fightin}, which reports which n-grams are more likely to be uttered by one group over the other given a prior distribution of words; currently, the priors are defined based on the provided dataset, however we hope to flexibly handle any user-provided priors in the future.
Below is an example of the log-odds analysis that shows the top 5 n-grams in the student's utterances over the tutor's.

\begin{figure}[h!]
    \centering
    \includegraphics[width=\linewidth]{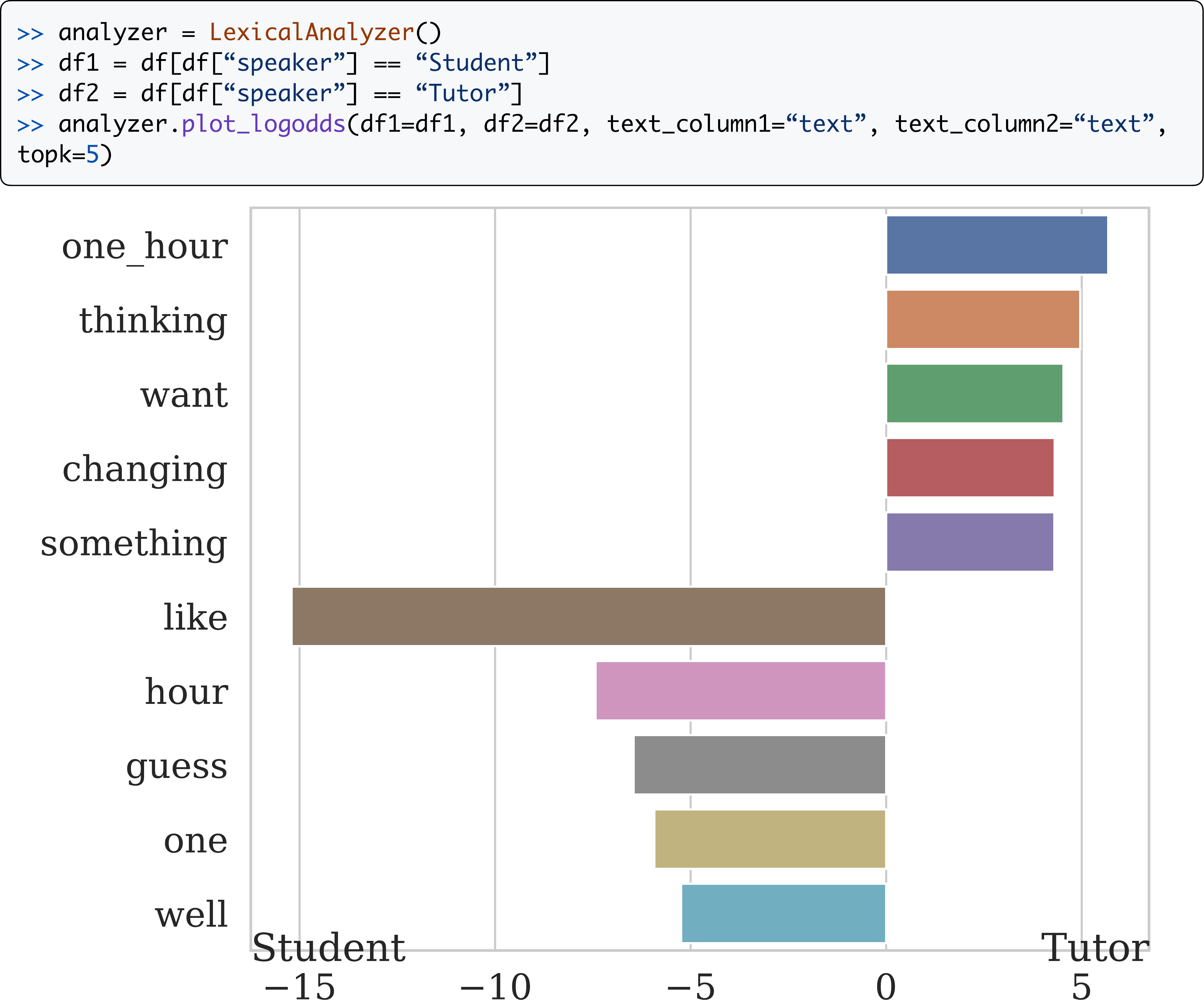}
\end{figure}

\newpage
\paragraph{\texttt{TemporalAnalyzer}.}
This module provides a time analysis of the annotations over the course of the conversation(s). 
Similar to \texttt{QuantitativeAnalyzer}, it can group and report the data in different ways. 
An important variable to this module is \texttt{num\_bins}, which indicates how many time bins the transcript should be split into; currently, the split is based on transcript lines, however we hope to support other split criteria in the future such as by word count.
Below is an example with speaker talk time.

\begin{figure}[h!]
    \centering
    \includegraphics[width=\linewidth]{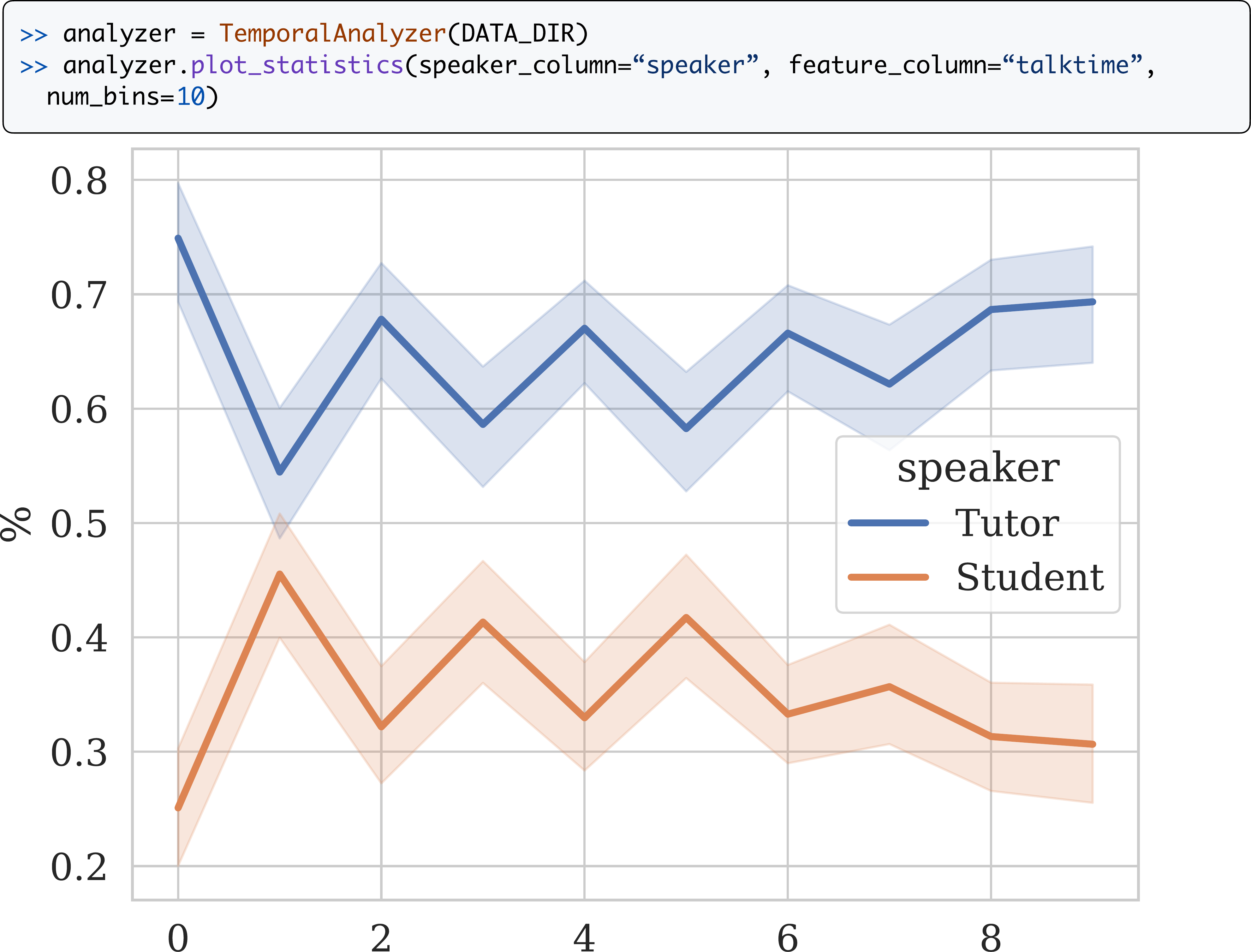}
\end{figure}

\paragraph{\texttt{GPTConversationAnalyzer}.}
This module uses GPT models accessible through the OpenAI API to analyze on the conversation-level with natural language.
Some prompts include summarizing the conversation (below example) or generating suggestions to the teacher/tutor on eliciting more student reasoning from \citet{wang-demszky-2023-chatgpt}.
The module has additional features (not shown) such as automatically truncating the transcript if it surpasses the model's context length, adding line numbers to the conversation or altering how the lines should be formatted.

\begin{figure}[h!]
    \centering
    \includegraphics[width=\linewidth]{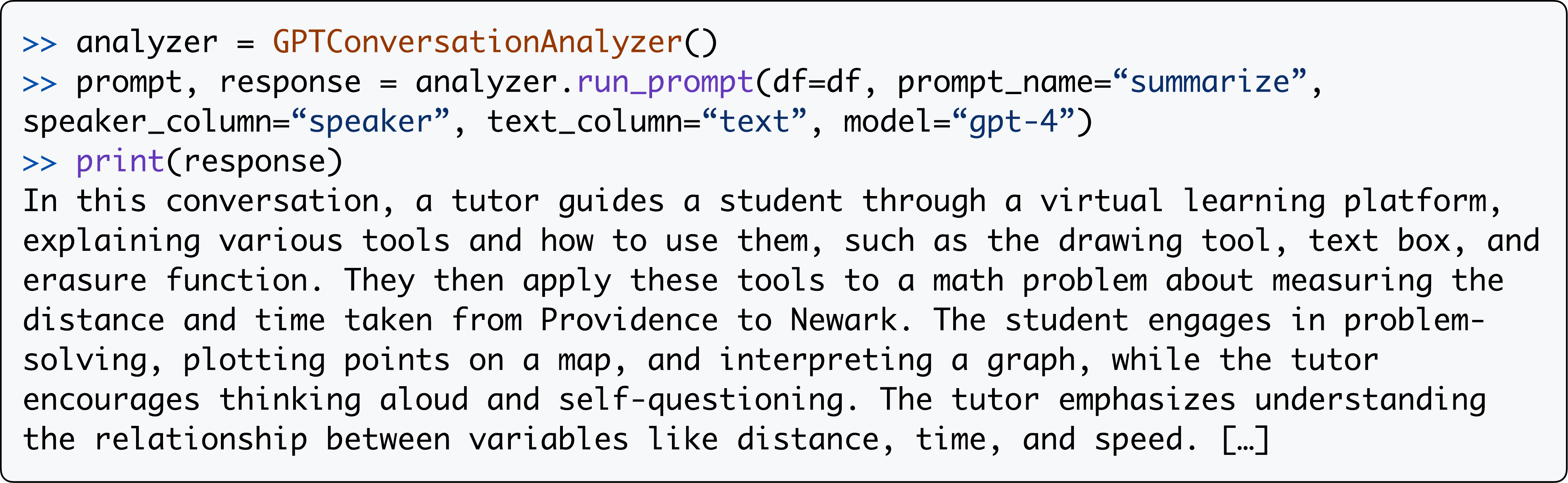}
\end{figure}

\section{Additional Resources: Basic Tutorials, Case Studies, and Paper Repository}
We create a suite of introductory tutorials and case studies of \toolname{} as Colab notebooks \href{https://github.com/stanfordnlp/edu-convokit/tree/main?tab=readme-ov-file#tutorials}{(link)}.
To demonstrate its wide applicability and generalizable design structure, we apply \toolname{} to three different education transcript datasets developed by different authors: 
NCTE, an elementary school classroom dataset \citep{demszky-hill-2023-ncte}; 
TalkMoves, a K-12 classroom dataset \citep{suresh2021using};
and Amber, a one-on-one 8th-9th grade tutoring dataset \citep{holt2023amber}.
For space reasons, we omit the findings of the case studies in this paper, but they can be found in our GitHub repository.
To centralize research efforts, we additionally contribute a paper repository that include papers that have used \toolname{} or have features incorporated into \toolname{}
\href{https://github.com/stanfordnlp/edu-convokit/blob/main/papers.md}{(link)}.

\section{Conclusion}

We introduce \toolname{}, an open-source library designed to democratize and enhance the study of education conversation data. 
Implemented in Python and easily accessible via GitHub and pip installation, it offers a user-friendly interface complete with extensive documentation, tutorials, applications to three diverse education datasets, and paper repository resource.
Based on extensive research experience, it incorporates best practices for pre-processing data and a series of different annotation measures grounded in prior literature, such as measuring student reasoning and talk time. 
It additionally supports several analysis modules, such as temporal analyses (e.g., talk time ratios), lexical analyses (e.g., word usage) and GPT-powered analyses (e.g., summarization).
Fostering a collaborative environment through its open-source nature, \toolname{} and its resources unify research efforts in this exciting interdisciplinary field to improve teaching and learning.

\section{Limitations}

There are limitations to \toolname{} which we intend on addressing in future versions of the library.
Some of the current limitations include: \toolname{} does not support transcription; 
it does not support connecting the language analyses to metadata, such as demographic data or learning outcomes, such as in \citet{demszky-hill-2023-ncte}; 
it only supports English-focused annotation methods;
many of its annotation models were trained on elementary and middle school mathematics, so they may not generalize to other domains;
and \toolname{}'s de-identification method assumes the speakers are known.
There are other existing de-identification methods that do not assume knowledge of the speaker names (one of which is also implemented in \toolname) however these methods are known to have high false-negative and false-positive rates.

\section{Ethics Statement}

The intended use case for this toolkit is to further education research and improve teaching and learning outcomes through the use of NLP techniques.
\toolname{} is intended for research purposes only.
\toolname{} uses data from existing public datasets that acquired consent from parents and teachers when applicable; for example, the NCTE dataset from \citet{demszky-hill-2023-ncte} acquired consent from parents and teachers for their study (Harvard's IRB  \#17768), and for the de-identified data to be publicly shared.
As stewards of this library which builds on these datasets, we are committed to protecting the confidentiality of the individuals and ask users of our library to do the same. 
It is important to note that inferences drawn using \toolname{} may not necessarily reflect generalizable observations (e.g., the student reasoning model was trained on elementary school math, and may not yield correct insights when applied to high school math).
Therefore, the analysis results should be interpreted with caution.
Unacceptable use cases include any attempts to identify users or use the data for commercial gain.
We additionally recommend that researchers who do use our toolkit take steps to mitigate any risks or harms to individuals that may arise. 

\section*{Acknowledgements}
We are thankful to Chris Manning, Omar Khattab, Ali Malik, Jim Malamut, Lucy Li for their feedback on the work.
Additionally, we are thankful to Yann Hicke for their contributions to \toolname{}.

\bibliography{custom}

\appendix

\end{document}